\documentclass{article}

\PassOptionsToPackage{numbers, compress}{natbib}
\usepackage[final]{neurips_data_2024}
\usepackage[utf8]{inputenc} % allow utf-8 input
\usepackage[T1]{fontenc}    % use 8-bit T1 fonts
\usepackage[colorlinks=true, linkcolor=red, urlcolor=blue, citecolor=blue]{hyperref}
\usepackage{url}            % simple URL typesetting
\usepackage{booktabs}       % professional-quality tables
\usepackage{chngpage}
\usepackage{multirow}
\usepackage{amsfonts}       % blackboard math symbols
\usepackage{nicefrac}       % compact symbols for 1/2, etc.
\usepackage{microtype}      % microtypography
\usepackage{xcolor}         % colors
\usepackage{authblk}        % for author affiliation
\usepackage{comment}
\usepackage{graphicx}
\usepackage{makecell}
\usepackage{array}
\usepackage{siunitx}
\usepackage{caption}
\usepackage{tabularx}
\usepackage{amsmath}
\usepackage{cleveref}
\usepackage{subfigure}
\usepackage[T1]{fontenc}
\usepackage{float} % Add this to the preamble for 'H' specifier

\newcommand{\sithead}[1]{\text{\textbf{#1}}}

\title{BioTrove: A Large Curated Image Dataset Enabling AI for Biodiversity}

\author[1,*]{Chih-Hsuan Yang}
\author[2,*]{Ben Feuer}
\author[1]{Zaki Jubery}
\author[3]{Zi K. Deng}
\author[2]{Andre Nakkab}
\author[1]{Md Zahid Hasan}
\author[1]{Shivani Chiranjeevi}
\author[2]{Kelly Marshall}
\author[1]{Nirmal Baishnab}
\author[1]{Asheesh K Singh}
\author[1]{Arti Singh}
\author[1]{Soumik Sarkar}
\author[3]{Nirav Merchant}
\author[2]{Chinmay Hegde}
\author[1]{Baskar Ganapathysubramanian}

\affil[1]{Iowa State University, Ames, IA 50011, USA}
\affil[2]{New York University, New York, NY 10003, USA}
\affil[3]{University of Arizona, Tucson, AZ 85721, USA}
\affil[*]{Joint first authors}
\affil[ ]{Correspondence: \texttt{chinmay.h@nyu.edu}, \texttt{baskarg@iastate.edu}.}

\begin{document}

\maketitle

\begin{comment}
Notes by Bella (10/21):
We are going to add the 200M full superset of the iNaturalist Open Dataset. Hence, the content for the intro of the dataset, the treemap, figure 1, figure 2, and Figure 3 , and the comparison table needs to change.  
\end{comment}
\begin{abstract}

We introduce \textsc{BioTrove}, the largest publicly accessible dataset designed to advance AI applications in biodiversity. Curated from the iNaturalist platform and vetted to include only research-grade data, \textsc{BioTrove} contains 161.9 million images, offering unprecedented scale and diversity from three primary kingdoms: \textit{Animalia} ("animals"), \textit{Fungi} ("fungi"), and \textit{Plantae} ("plants"), spanning approximately 366.6K species. Each image is annotated with scientific names, taxonomic hierarchies, and common names, providing rich metadata to support accurate AI model development across diverse species and ecosystems.

We demonstrate the value of \textsc{BioTrove} by releasing a suite of CLIP models trained using a subset of 40 million captioned images, known as \textsc{BioTrove-Train}. This subset focuses on seven categories within the dataset that are underrepresented in standard image recognition models, selected for their critical role in biodiversity and agriculture: \textit{Aves} ("birds"), \textit{Arachnida} ("spiders/ticks/mites"), \textit{Insecta} ("insects"), \textit{Plantae} ("plants"), \textit{Fungi} ("fungi"), \textit{Mollusca} ("snails"), and \textit{Reptilia} ("snakes/lizards"). To support rigorous assessment, we introduce several new benchmarks and report model accuracy for zero-shot learning across life stages, rare species, confounding species, and multiple taxonomic levels.

We anticipate that \textsc{BioTrove} will spur the development of AI models capable of supporting digital tools for pest control, crop monitoring, biodiversity assessment, and environmental conservation. These advancements are crucial for ensuring food security, preserving ecosystems, and mitigating the impacts of climate change. \textsc{BioTrove} is publicly available, easily accessible, and ready for immediate use.

\end{abstract}

%In this paper, we introduce BioTrove, the largest publicly accessible dataset designed to advance AI for biodiversity. BioTrove contains 126 million images, surpassing existing datasets in scale and diversity, and includes image-language paired data for various species like birds, arachnids, insects, plants, fungi, mollusks, and reptiles. Each image is annotated with scientific names, taxonomic details, and common names, facilitating robust AI model training.

%We developed the \textsc{BioTrove-CLIP}-40M model using a subset of 40 million images and created benchmark datasets for rigorous evaluation, reporting accuracy for zero-shot learning and evaluations across life stages. BioTrove enables AI model development for pest control, crop monitoring, and biodiversity assessment, contributing to food security, ecosystem preservation, and climate change mitigation. The dataset is publicly available and ready for immediate use.

\section{Introduction}
\vspace{-0.15in}
AI advances are poised to play a crucial role in biodiversity conservation, ecology management, and agriculture. Already, AI tools have been shown to enable automated species identification, monitoring of ecological changes, and optimization of crop management \citep{shivaprakash2022potential, chiu2020agriculture}. However, standard AI approaches for biodiversity applications persistently face major challenges. Training datasets are labor-intensive and costly to create; they cover only a narrow set of visual concepts; standard vision models excel at single tasks but require extensive retraining for new tasks; models often struggle with generalizing to unseen labels and new environments, limiting their effectiveness in real-world applications \citep{roy2024towards, guldenring2021self}.  Models that perform well on benchmarks often fail in the wild \citep{geirhos2018imagenet, alcorn2019strike}. Standard computer vision datasets (ImageNet and its successors) have significant limitations, including incorrectly labeled images, geographical and cultural biases, and overlapping or ill-defined labels, all of which impair the development of high-performant AI models \citep{luccioni2023bugs}. Consequently, there is a critical need for large, diverse, accurately annotated datasets that are specific to biodiversity, ecology, and agricultural research \citep{muller2023soundscapes, lu2022generative}. 

In response to this need, several datasets have been introduced. Perhaps the most well-known (raw) pool of biodiversity images on the Web is iNaturalist~\cite{van2018inaturalist}, from which several curated datasets have been sourced, among them being iNat2021~\cite{unger2021inaturalist} with 2.7M images of over 10,000 species of plants, animals, and fungi. However, insects (which comprise a very large fraction of extant species) are under-represented in this dataset. IP102~\citep{Wu2019Insect}, Insecta~\cite{feuer2024insect}, and the more recent \textsc{BioScan-1M}~\cite{gharaee2024step}, are alternative datasets that focus on the Insecta Class. Perhaps the latest advance in such research is \textsc{TreeOfLife-10M}~\cite{stevens2023bioclip}, which is currently the state-of-the-art dataset of text-annotated biological images, comprising 10M images with approximately 450K unique taxonomic classes. 

\begin{figure}[!t]
    \centering
    \includegraphics[width=\textwidth]{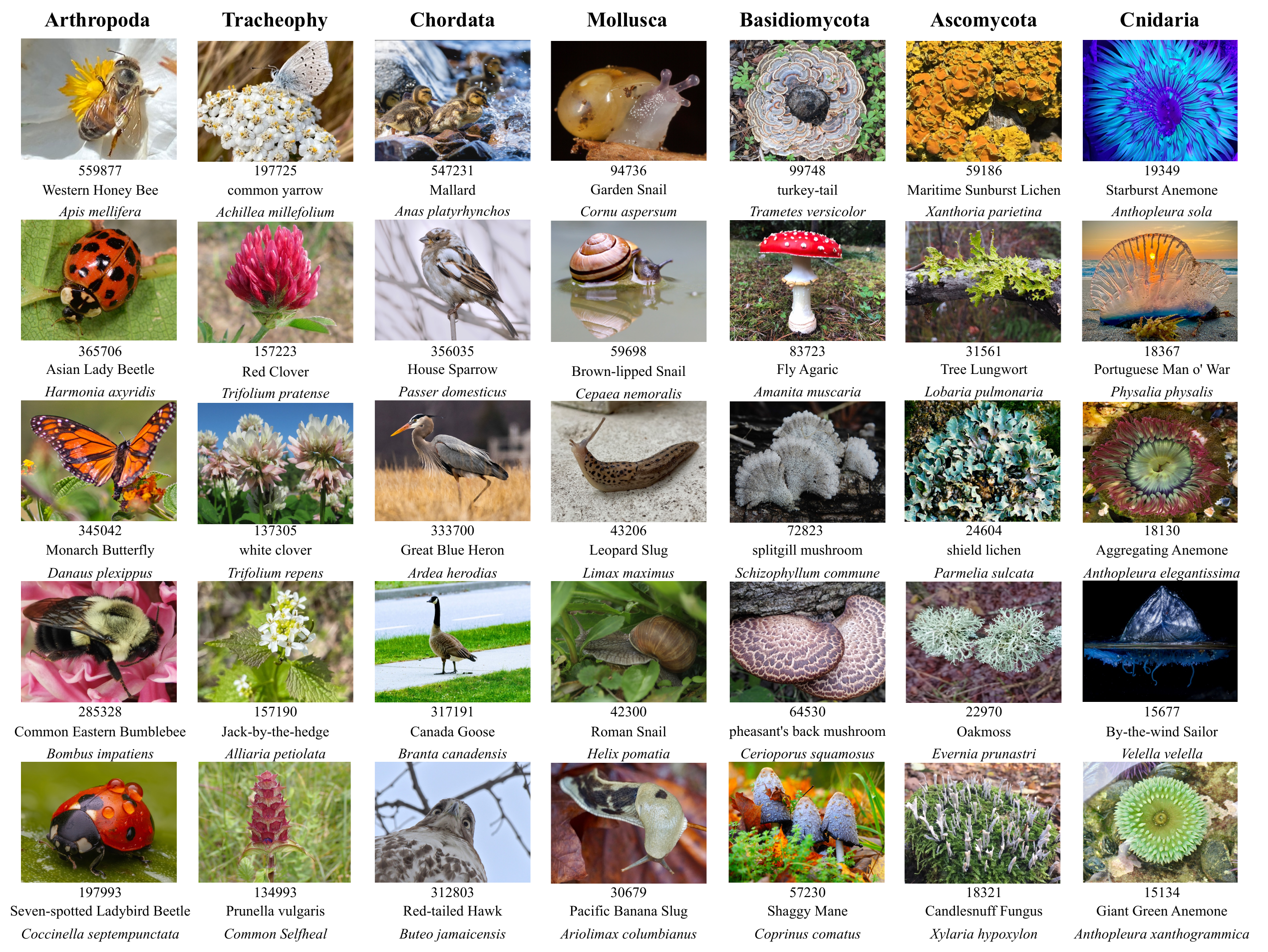}
    \caption{\sl \textbf{Top Seven Phyla in the \textsc{BioTrove} Dataset.} This figure displays the seven most frequently occurring phyla within \textsc{BioTrove}, which is curated to include data exclusively from the three primary kingdoms: \textit{Animalia}, \textit{Plantae}, and \textit{Fungi}. For each phylum, the five most common species are shown, including their scientific names, common names, and the number of images per species. The phyla are ordered by species diversity, with the most diverse phylum on the right and the least diverse on the left.}\vspace{-0.15in}
    \label{fig:7_categories}
\end{figure}

In this paper, we contribute to advancing biodiversity AI research by curating and releasing \textbf{\textsc{BioTrove}}, a dataset comprising \textbf{161.9 million captioned images} across approximately \textbf{366.6K species}. This dataset surpasses all previous collections in both scale and diversity, representing the largest publicly available, ``AI-ready" dataset of curated biodiversity images. Each image in \textsc{BioTrove} is paired with language data and spans a diverse range of taxonomic groups, including \textit{Reptilia} (reptiles), \textit{Plantae} (plants), \textit{Mollusca} (mollusks), \textit{Mammalia} (mammals), \textit{Insecta} (insects), \textit{Fungi} (fungi), \textit{Aves} (birds), \textit{Arachnida} (arachnids), \textit{Animalia} (animals), \textit{Amphibia} (amphibians), and \textit{Actinopterygii} (ray-finned fish). The dataset spans global regions, supporting robust training across diverse environmental contexts. Representative examples are shown in Figure~\ref{fig:7_categories}, and additional details are provided on the project~\href{https://baskargroup.github.io/BioTrove/}{website}.

Each image in \textsc{BioTrove} originates from the iNaturalist community science platform~\cite{van2018inaturalist} and is annotated with detailed metadata, including the common name, scientific name, and complete taxonomic hierarchy. This curated metadata provides research-grade high-quality text annotations that enhance AI model training. Additionally, we open-source a data management pipeline, \textsc{BioTrove-Process}, to facilitate interaction with \textsc{BioTrove} metadata. With \textsc{BioTrove-Process}, researchers can efficiently filter and balance data by selecting specific taxonomic categories, adjusting for taxonomy level, and managing species distribution to reduce skewness. This enables users to create custom subsets that align with their research goals while maintaining consistency in species representation.

To showcase the capabilities of \textsc{BioTrove}, we introduce two technical contributions. First, we train and release \textsc{BioTrove-CLIP}, a suite of vision-language foundation models, using a subset, \textsc{BioTrove-Train}, consisting of approximately 40M images from \textsc{BioTrove} and representing around 33K species. This subset, constructed with \textsc{BioTrove-Process}, includes diverse taxa, specifically focusing on birds (\textit{Aves}), spiders/ticks/mites (\textit{Arachnida}), insects (\textit{Insecta}), plants (\textit{Plantae}), fungi (\textit{Fungi}), snails (\textit{Mollusca}), and snakes/lizards (\textit{Reptilia}). These taxonomic classes were selected to capture a broad range of species—outside of charismatic megafauna—that critically impact biodiversity. The models exhibit robust generalization capabilities, demonstrating high zero-shot and few-shot performance on unseen taxa when using either common or scientific names. We anticipate that \textsc{BioTrove-CLIP} will serve as a valuable foundation for biodiversity-related applications and can be further fine-tuned for specific research needs.

Second, we rigorously quantify the performance of our foundation models on five existing fine-grained image classification benchmarks, as well as on three newly curated test datasets. We find that \textsc{BioTrove-CLIP} models comfortably achieve the state-of-the-art in certain settings, while both the original (OpenAI) \textsc{CLIP} model as well as \textsc{BioCLIP}~\cite{stevens2023bioclip}  excel in certain other settings. We analyze these findings in further detail below, but overall we hope that our dataset can be used by the AI community as a testbed for further algorithmic and scaling research in fine-grained image recognition.

\begin{figure}[!t]
    \centering
    \includegraphics[width=\textwidth]{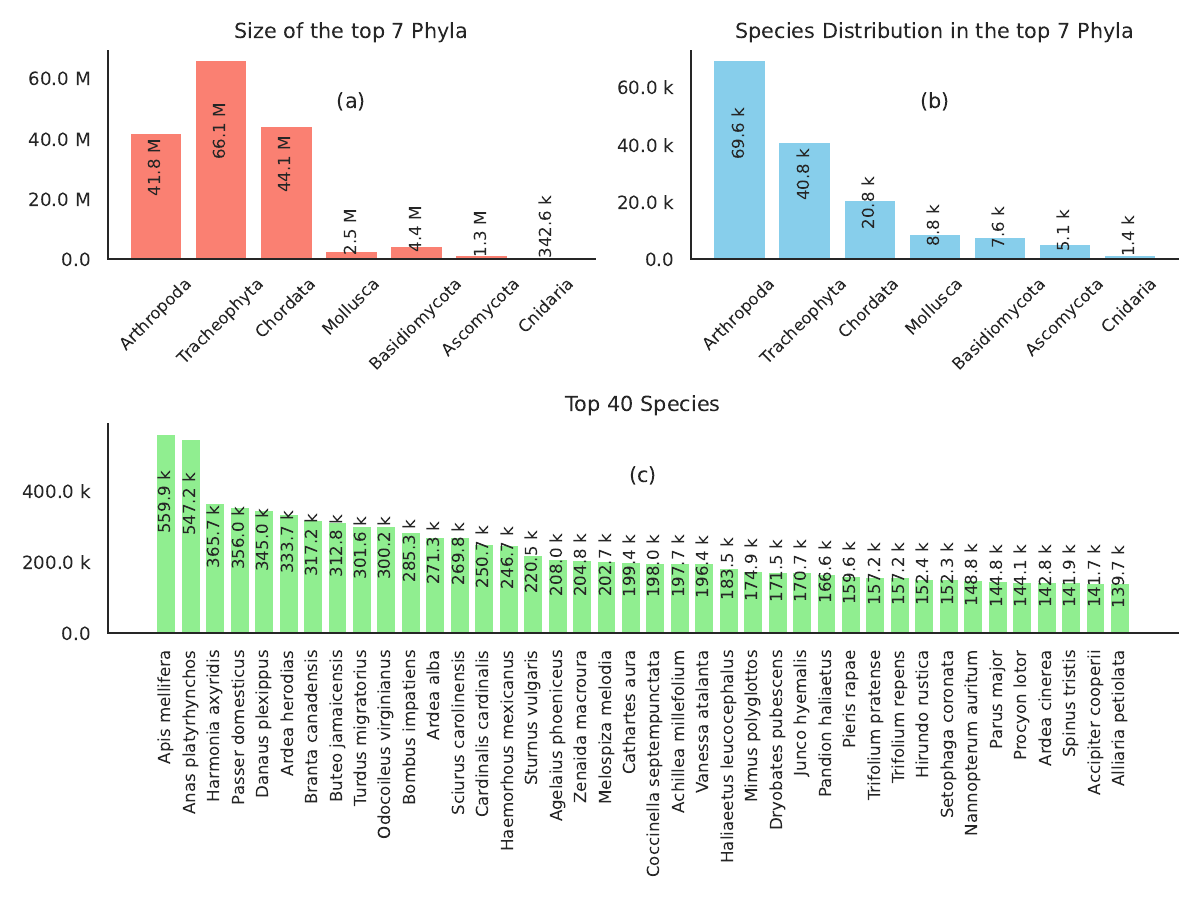}
    \caption{\sl \textbf{Distribution of the BioTrove dataset.} (a) Size of the top seven Phyla in the BioTrove dataset. (b) Species counts for the top seven Phyla. (c) The 40 highest occurring species in entire BioTrove dataset.}
    \vspace{-0.1in}
\label{fig:combined_distribution}
\end{figure}

The remainder of this paper is organized as follows. Section~\ref{sec:BioTrove} introduces the \textsc{BioTrove} dataset, the dataset's salient characteristics, and a comparison with previous work. Section~\ref{sec:workflow} details our curation methodology. Section~\ref{sec:benchmarks} introduces our newly proposed test datasets and their characteristics. Section~\ref{sec:experiments} details our new \textsc{BioTrove-CLIP} models and their benchmark performance relative to previous work. Section~\ref{sec:future} concludes with a discussion of limitations and potential future directions. 

\vspace{-0.1in}
\section{The BioTrove Dataset}
\label{sec:BioTrove}
\vspace{-0.1in}
%Our central contribution in this paper is the \textsc{BioTrove} dataset with the following salient features.

\begin{figure}[!t]
    \centering
        \includegraphics[width=\textwidth]{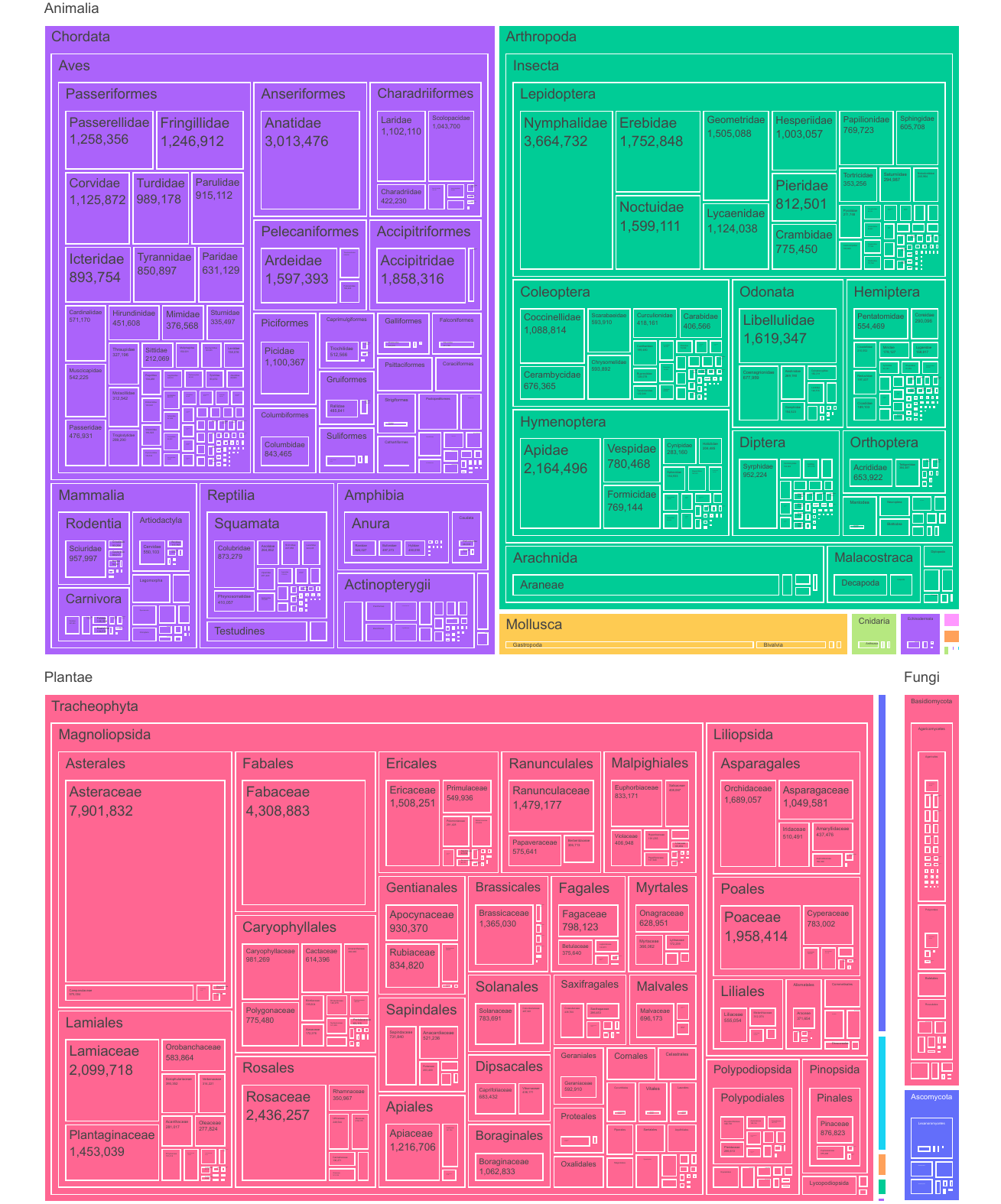}
        \caption{\sl \textbf{Treemap diagram of the BioTrove dataset}, starting from \emph{Kingdom}. The nested boxes represent phyla, (taxonomic) classes, orders, and families. Box size represents the relative number of samples.}
        \vspace{-0.1in}
        \label{fig:treemap}
\end{figure}

%\subsection{Characteristics}

\paragraph{Characteristics.} \textsc{BioTrove} comprises over 161.9 million images spanning 372,966 species. This dataset is an order of magnitude larger than existing biodiversity datasets, such as the state-of-the-art \textsc{TreeOfLife-10M} dataset, which it surpasses in \emph{scale} by a factor of nearly $13.5\times$ while maintaining comparable \emph{species diversity}. Figure~\ref{fig:7_categories} shows representative image samples, while Figure~\ref{fig:combined_distribution} displays the distribution of samples across the seven major categories with the most frequently observed species. Additionally, Figure~\ref{fig:treemap} illustrates the range of phyla, taxonomic classes, orders, and families represented in the dataset.

\textsc{BioTrove} includes only research-grade data and publicly accessible licensed content for research purposes from iNaturalist, which designates observations as research-grade once they meet strict validation criteria. To qualify, two or more experienced iNaturalist community members—naturalists, biologists, or citizen scientists—must agree on the species identification. Additionally, the observation must meet other requirements, such as a clear photograph and precise geolocation data. Recent experiments have shown that iNaturalist’s Research Grade observations achieve approximately 97\% accuracy, underscoring the reliability of this community-based validation process~\cite{inaturalist_accuracy}. Furthermore, iNaturalist continuously enhances data quality by refining validator criteria and implementing new data quality assessment measures, ensuring \textsc{BioTrove} remains a robust dataset for scientific use.

Each image sample in \textsc{BioTrove} is enriched with detailed, curated metadata that facilitates efficient filtering by species count and taxonomic information. The metadata includes common names, scientific names, and hierarchical taxonomic data, which enhances the usability of the dataset for AI model training. For the complete list of metadata fields, see Table~\ref{tab:annotation_details}.

%\ref{tab:annotation_details}. 
%This comprehensive approach ensures researchers can start their work quickly, focusing on model training and evaluation without the need to preprocess the data themselves.

Along with the dataset, we also release our data curation tooling pipeline: \textsc{BioTrove-Process}, which enables users to easily access and manipulate the dataset. This pipeline allows researchers to select specific categories across different taxonomic levels, visualize data distributions, and effectively manage class imbalance according to their needs. It facilitates the downloading of specific images by their URLs and provides image-text pairs as well as user-defined chunks to support various AI applications. \textsc{BioTrove-Process} thus enables users to define custom subsets of \textsc{BioTrove} with ease, making the dataset fully AI-ready and reducing barriers to follow-up research in biodiversity-focused AI.

%For more detailed analysis and figures, please refer to Figures \ref{fig:treemap}, \ref{fig:7_categories}, and \ref{fig:combined_distribution}.

%\subsection{Dual-Language Text Descriptions}
\paragraph{Dual-language text descriptions.} We adopt both common and scientific names since Latin is a low-resource language, and current AI models do not perform well on scientific names alone in a zero-shot manner. We found that a well-structured text description that integrates common names, scientific names, and detailed taxonomic hierarchies facilitates the learning of relationships between Latin and English terms, thereby improving the models' applicability in scientific contexts \citep{de2024unsupervised, sprugnoli2020building, wijayanti2023learning}. Moreover, incorporating the taxonomic hierarchy enables models to more effectively associate visual data with taxonomic terminology \citep{mars2022word, chen2021low}. This matches the guidelines suggested by \textsc{BioCLIP} \cite{stevens2023bioclip}  to enhance model performance and generalization.
\begin{table}[!t]
    \centering
    \caption{\sl \textbf{Annotations provided in the BioTrove Dataset.}     \label{tab:annotation_details}}
    \begin{tabular}{ll}
        \toprule
        Text Type & Description \\
        \midrule
        Common Name & Vernacular name (e.g., Western honey bee) \\
        Scientific Name & Genus and species (e.g., \textit{Apis mellifera}) \\
        Taxonomic Name & Flattened taxonomy concatenated into a single string \\
        Taxonomic Rank & Specific level in the hierarchy (e.g., subspecies, species) \\
        
        \bottomrule
    \end{tabular}
    \vspace{-0.1in}
\end{table}
%\noindent The inclusion of both common and scientific names at various taxonomic levels improves the model's ability to learn and generalize across different linguistic representations. This dual-language strategy is essential for enhancing the robustness and applicability of AI models in biodiversity and ecological research. By following the recommendations from the BioCLIP paper, our dataset ensures high-quality annotations that support accurate and efficient species identification and classification \citep{stevens2023bioclip}.
%\subsection{Content Considerations and Privacy Measures}
%\label{Content_Considerations_and_Privacy_Measures}
\textbf{Privacy Measures:} The images of \textsc{BioTrove} were sourced from the iNaturalist Open Dataset, whose metadata included Personally Identifiable Information (PII). This included information about observers, such as their usernames and sometimes their real names if they have chosen to share that information publicly. We removed all such fields to ensure that no PII is present in the metadata associated with \textsc{BioTrove} samples, ensuring the privacy of all contributors.
\textbf{License:} During curation, we took care to include only images from iNaturalist Open Data, which are all licensed under either the \texttt{CC0}, or \texttt{CC-BY}, or \texttt{CC-BY-NC} licenses. This ensures that all our images are available for public research purposes. 
\textbf{Offensive Content:} Some of our URLs may point to images that users could find disturbing or harmful, such as photos of dead or dismembered animals. We retained these types of images since they sometimes can provide valuable scientific data about wildlife, including information on predation events, roadkill, and other occurrences relevant to conservation and biodiversity studies. Although iNaturalist relies on user contributions and community moderation to maintain the quality and appropriateness of the data, we acknowledge that the vast and diverse nature of the data means that some offensive or inappropriate content might be present.

%\subsection{Comparisons with the State of the Art}

%I would like to express my deepest appreciation and admiration for researchers' significant contributions to the development of datasets for AI in biodiversity. Therefore, I would like to bring up and compare our work with the two previous outstanding datasets, the TreeOfLife and BioScan~\cite{gharaee2024step} datasets, the latter of which was featured in the 2023 NeurIPS Dataset and Benchmark paper.

%The BioTrove, TreeOfLife, and BioScan \cite{gharaee2024step} datasets are pivotal resources for AI applications in biodiversity, agriculture, and ecological research. BioTrove, the largest and most comprehensive of the three datasets with 161.9 million images, is derived from the iNaturalist Open Dataset and provides dual-language labels (common and scientific names) along with detailed taxonomic hierarchies. 
%It includes a ready-to-use pipeline for data processing and efficient algorithms for managing class imbalance, making it the largest and easiest-to-use AI dataset for biodiversity. Notably, the BioTrove-CLIP model is trained from a subset called BioTrove-Train. 
Our closest comparisons are with \textsc{BioScan-1M} (which appeared in NeurIPS 2023 Datasets and Benchmarks) and \textsc{TreeOfLife-10M} (which will appear in CVPR 2024). \textsc{BioScan-1M} focuses solely on the Insecta Class and provides scientific names, taxonomic ranks, as well as DNA barcodes. The \textsc{TreeOfLife-10M} dataset comprises 10.4 million images, integrating data from iNat2021~\cite{unger2021inaturalist}, \textsc{BioScan-1M}, and a fresh set of image samples sourced from the Encyclopedia of Life (EOL). It also supports dual-language labels and detailed taxonomic hierarchies and was used to train the \textsc{BioCLIP} vision-language model. 
See Table~\ref{tab:dataset_comparison_similar_dataset_compare} for essential differences.

\begin{table}[!b]
    \centering
    \caption{\sl \textbf{Comparison} of {BioTrove} with existing biodiversity datasets. \label{tab:dataset_comparison_similar_dataset_compare}}
    \resizebox{\textwidth}{!}{
    \renewcommand{\arraystretch}{1.2} % Add space between rows
    \begin{tabular}{|>{\centering\arraybackslash}m{2.5cm}|>{\centering\arraybackslash}m{3.5cm}|>{\centering\arraybackslash}m{3.5cm}|>{\centering\arraybackslash}m{3.5cm}|}
        \hline
        \textbf{Feature} & \textbf{BioTrove} & \textbf{TreeOfLife} & \textbf{BioScan} \\ \hline
        \textbf{Size} & 161.9 million images & 10.4 million images & 1.1 million images \\ \hline
        \textbf{Diversity} & 366.6K species & 454.1K species & 8.3K \\ \hline
        \textbf{Labels Provided} & Dual language (common and scientific names), detailed taxonomic hierarchies & Dual language (common and scientific names), detailed taxonomic hierarchies & Single language (scientific names), taxonomic ranks (family to species), DNA barcodes \\ \hline
        \textbf{Data Source} & iNaturalist Open Dataset & iNaturalist, Encyclopedia of Life (EOL), \textsc{BioScan}-1M & Specimens from Malaise traps, DNA barcodes matched to BOLD \\ \hline
        \textbf{\makecell{Key \\ Features}} & Ready-to-use pipeline, reduce class imbalance, high-quality annotations, supports \textsc{BioTrove-CLIP} & Rich hierarchical representations, comprehensive metadata, supports \textsc{BioCLIP} & Focus on insects, high-resolution images, detailed taxonomic annotation, DNA codes \\ \hline
    \end{tabular}
    }
\end{table}

 \vspace{-0.15in}
\section{Data Collection and Curation Methodology}
\label{sec:workflow}
\vspace{-0.1in}
%\subsection{iNaturalist Open Dataset and AI Usability}

\paragraph{Challenges with iNaturalist Open Data.} All of \textsc{BioTrove} is sourced from the iNaturalist Open Data community science platform, which (in all) comprises over 280M  biodiversity-relevant observations shared by users. However, there are still significant gaps in usability for AI research. The photos and metadata, although easily downloadable, are provided in four separate metadata sheets that are not ready to use. Taxa information is encoded as numerical IDs, requiring additional API calls and non-trivial lookups to convert these into common or scientific names. The multiple metadata sheets structure is fragmented across four separate files—photos, taxa, observations, and observers—adding complexity to data integration. Managing data balance and filtering out species with too few images can lead to biases toward common (charismatic) species and an imbalanced training process. 

% \citep{buda2018systematic,ali2019imbalance,belkin2018overfitting}. Additionally, the dataset's sheer size, with approximately 300 million images, makes it difficult to work with subsets without handling the entire dataset, compounded by the rate limitations of the iNaturalist API. This can lead to biases towards common species, loss of biodiversity representation, overfitting, underrepresented classes, and an imbalanced training process \citep{thabtah2020data,fernandez2011addressing,krawczyk2016learning}.

\paragraph{Curation of \textsc{BioTrove}.} The iNaturalist Open Dataset comprises a collection of 284.2 million images stored on an AWS S3 bucket as of 2024-09-27, with associated metadata provided across four separate CSV files (\texttt{photos}, \texttt{observations}, \texttt{taxa}, and \texttt{observers}). Details on each of these files are presented in Section~\ref{sec:inaturalist} in the Appendix. Although these files contain a wealth of valuable information, they are structured for rapid retrieval rather than AI-readiness. To address this, we curate the metadata into a streamlined, AI-optimized format.

We populate an SQL database with each CSV file as an individual SQL table, then create an aggregated table by joining \texttt{photos}, \texttt{observations}, and \texttt{taxa} on their relational columns, discarding irrelevant columns.  In this aggregate table, we add a new column populated with the Amazon S3 URL and generate individual columns for taxonomic kingdom, phylum, class, order, family, genus, and species. 

\textsc{BioTrove} includes only research-grade images from the \textit{Animalia}, \textit{Plantae}, and \textit{Fungi} kingdoms, filtering out other domains to maintain a clear biodiversity focus. To achieve this filtering, we apply strict taxonomic criteria, ensuring only these three kingdoms are represented. The iNaturalist metadata files lack common names, so we reconstruct this information by cross-referencing species names from the iNaturalist Taxonomy DarwinCore Archive, updated monthly. This enriched metadata, including common names, is then appended to the SQL table. The final curated dataset is exported as parquet files, available for public access on \href{https://huggingface.co/datasets/BGLab/BioTrove}{HuggingFace}.

\paragraph{Data Filtering and Preprocessing.} As outlined, \textsc{BioTrove} includes structured metadata that is both comprehensive and easy to work with, featuring full taxonomic information and direct URLs to image files. To further support accessibility, we release an accompanying software pipeline that allows users to filter specific categories, visualize data distributions, and manage dataset imbalances effectively. These tools make it simple for researchers to interact with \textsc{BioTrove}, creating tailored subsets based on their specific needs. The iNaturalist data, sourced from citizen science contributions, has inherent variability in species representation, with some species documented extensively and others less so. To address this, our tools enable user-defined filters to exclude species with fewer than a set number of images and to cap image counts per species, thus supporting more balanced model training.

To further mitigate dataset imbalances (detailed in our experiments section), we use a semi-global shuffling strategy in which the data is organized into chunked tar files. These files are shuffled, divided into smaller groups, and then merged into larger batches to ensure a balanced species distribution within each batch. This method enhances dataset integrity, helping to prevent the overrepresentation of any single species across the batches.

 \vspace{-0.15in}
\section{Models and Benchmarks}
\label{sec:benchmarks}
\vspace{-0.1in}
We now showcase and demonstrate the utility of the \textsc{BioTrove} dataset by creating and benchmarking \textsc{BioTroveCLIP}, a new suite of vision-language foundation models for biodiversity. 

 \vspace{-0.1in}
\subsection{BioTrove-Train}
\label{BioTrove-Train}
\vspace{-0.1in}
\textsc{BioTrove-Train} is a curated subset comprising approximately 40M samples and 33K species, focused specifically on seven taxonomic categories: \textit{Aves}, \textit{Arachnida}, \textit{Insecta}, \textit{Plantae}, \textit{Fungi}, \textit{Mollusca}, and \textit{Reptilia}. As discussed previously, the \textsc{BioTrove} dataset is accompanied by a flexible pipeline that enables users to apply customized filtering to select specific categories or subsets based on research needs, thereby allowing researchers to generate their own training datasets. For \textsc{BioTrove-Train}, these seven categories were pre-selected due to their significant impact on biodiversity and agricultural ecosystems, as well as their relative underrepresentation in standard image recognition models. Unlike megafauna, which are typically well-represented in existing models, these categories address unique challenges in biodiversity-focused AI.

This subset comprises data posted on iNaturalist prior to 2024-01-27. We applied strict filtering criteria to ensure high-quality data, excluding species with fewer than 30 images and capping the maximum number of images per species at 50,000. To maintain balance, we employed a semi-global shuffling method, organizing the data into mini-batches of approximately 50,000 samples. From these, 95\% were randomly selected for training and validation, while the remaining 5\% were reserved for testing. Detailed statistics can be found in Table~\ref{tab:BioTrove-Train}.

\begin{table}[!ht]
    \centering
    \caption{\sl \textbf{Training data sources used in \textsc{BioTrove-Train} and Diversity in Different Taxonomy Levels}. We integrate taxonomic labels into the images.     \label{tab:BioTrove-Train}}
    \resizebox{0.9\textwidth}{!}{
    \begin{tabularx}{\textwidth}{@{}p{10cm} p{3.5cm}@{}}
        % Left Table
        \begin{tabular}{@{}>{\centering\arraybackslash}m{1.9cm} >{\centering\arraybackslash}m{4.2cm} >{\centering\arraybackslash}m{1.2cm} >{\centering\arraybackslash}m{1.2cm}@{}}
            \toprule
            \textbf{Dataset} & \textbf{Description} & \textbf{Images} & \textbf{Unique Classes} \\
            \midrule
            \textsc{TreeOfLife-10M} & Dataset combines a subset of iNaturalist, Encyclopedia of Life (EOL), \textsc{BioScan}-1M. & 10.4M & 454,103 \\
            \textsc{BioTrove-Train} & One subset of BioTrove with size 40M. & 39.9M & 33,364 \\
            \bottomrule
        \end{tabular}
        &
        % Right Table
        \begin{tabular}{@{}>{\centering\arraybackslash}m{1.75cm} >{\centering\arraybackslash}m{1.75cm} >{\centering\arraybackslash}m{1.75cm}@{}}
            \toprule
            \textbf{Level} & \textbf{Uniques} \\
            \midrule
            kingdom & 3 \\
            phylum & 14 \\
            class & 50 \\
            order & 311 \\
            family & 1692 \\
            genus & 11506 \\
            species & 33364 \\
            \bottomrule
        \end{tabular}
    \end{tabularx}
    }
\end{table}
 \vspace{-0.1in}

\subsection{New Benchmarks}
\label{sec:BioTrove-benchmarks}
\vspace{-0.1in}
We created three new benchmark datasets, all of which are non-overlapping curated subsets of the \textsc{BioTrove} dataset. These benchmarks focus on fine-grained image classification within the seven taxonomic categories: \textit{Aves}, \textit{Arachnida}, \textit{Insecta}, \textit{Plantae}, \textit{Fungi}, \textit{Mollusca}, and \textit{Reptilia}. All benchmarks presented here are independent and strictly within these seven categories, without overlapping with each other or with the \textsc{BioTrove-Train} subset. Additionally, we report results on several established benchmarks from the literature (see Table~\ref{tab:eval-data}).

\noindent\textbf{BioTrove-Balanced.} To ensure balanced species representation across the seven key taxonomic categories, we curate the \textsc{BioTrove-Balanced} benchmark. Each category includes up to 500 species, with 50 images per species, resulting in a total of 112,209 images. This balanced dataset provides a consistent foundation for model performance evaluations. The exact species counts for each category are detailed in Table~\ref{tab:BioTrove-Train-Categories-species-count} (see Appendix).

\noindent\textbf{BioTrove-Unseen.} To assess the ability of models to generalize to previously unseen species within the seven categories, we curated the \textsc{BioTrove-Unseen} benchmark. This dataset includes species from \textsc{BioTrove-Train} with fewer than 30 instances, ensuring they were unseen during training. Each species is represented by at least 10 images, with a total of 11,983 images. This benchmark tests the models' robustness on rare species not encountered during training.

\begin{figure}[htbp]
\centering    
\includegraphics[width=\textwidth]{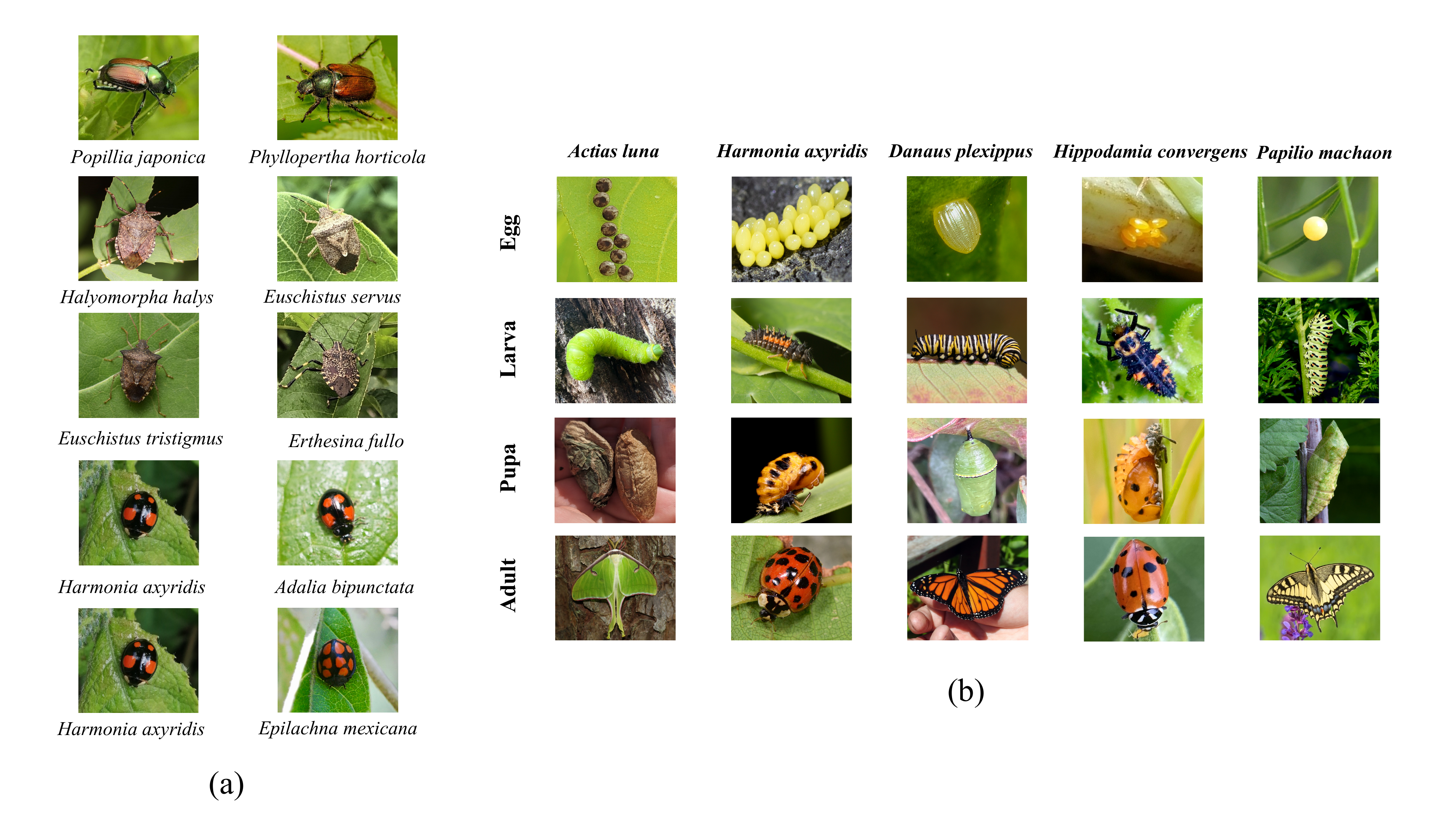}
\caption{\sl (a) Example images from BioTrove-Unseen. (b) \textsc{BioTrove-Life-Stages} with 20 class labels: four life stages (egg, larva, pupa, and adult) for five distinct insect species.}
\label{fig:comp_life_stages_benchmark}
\end{figure}

\noindent\textbf{BioTrove-LifeStages.} The \textsc{BioTrove-LifeStages} benchmark evaluates the model’s ability to recognize species across different developmental stages, focusing on insect species that exhibit significant visual variations throughout their life cycle. This dataset contains 20 labels representing four life stages (egg, larva, pupa, and adult) for five distinct insect species. The data was collected via the observation export feature on the iNaturalist platform between February 1, 2024, and May 20, 2024, ensuring no overlap with the training dataset. This benchmark allows for comprehensive evaluations of model performance across various life stages (see Figure~\ref{fig:comp_life_stages_benchmark}).

\begin{comment}
\begin{figure}[ht]
    \centering
   \includegraphics[width=0.85\textwidth]{figures/lifestages.pdf}
    \caption{Life stages of benchmark species }
    \label{fig:lifestages}
\end{figure}

\begin{figure}[ht]
    \centering
   \includegraphics[width=0.5\textwidth]{figures/hierarchical_distribution_treemap.pdf}
    \caption{Hierarchical Distribution of Taxonomic Levels (draft)}
    \label{fig:treemap}
\end{figure}

\begin{figure}[ht]
    \centering
   \includegraphics[width=0.75\textwidth]{figures/distribution_before_after_one_batch.pdf}
    \caption{Zahid: Species distribution in a training batch before and after semi-global shuffling (draft)}
    \label{fig:semi-global}
\end{figure}

\end{comment}

 \vspace{-0.15in}
\subsection{BioTrove-CLIP: New vision-language foundation models for biodiversity}
\label{training_details}
\vspace{-0.1in}
We use \textsc{BioTrove-Train} to train new CLIP-style foundation models and then evaluate them on zero-shot image classification tasks. Following the implementation of \citet{stevens2023bioclip}, we utilize a ViT-B/16 architecture initialized from the OpenAI CLIP weights~\citep{radford2021learning}, and train for 40 epochs. We also train a ViT-L/14 model from the MetaCLIP~\cite{xu2023metaclip} checkpoint for 12 epochs and a ViT-B/16 from the BioCLIP checkpoint for 8 epochs. All training hyperparameters are included in the Appendix (Section \ref{sec:deets}). We compare with OpenAI's ViT-B/16 CLIP model, the BioCLIP ViT-B/16 checkpoint, and MetaCLIP-CC ViT-L/14. 
%; we find that the latter, which was trained on more data and utilizes a larger ViT model, generally outperformed OpenAI's ViT-B/16. 
We publicly release all code needed to reproduce our results \href{https://github.com/baskargroup/BioTrove/}{here}.

 \vspace{-0.15in}
\section{Experimental Results}
\label{sec:experiments}
\vspace{-0.1in}

%\noindent\textbf{Benchmark Datasets.} 

\begin{table*}[t]
    \centering
    \footnotesize
    \caption{
        Existing benchmark datasets; our novel datasets are described separately in \cref{sec:BioTrove-benchmarks}.
        \label{tab:eval-data}
    }
    \setlength\tabcolsep{3pt}
    \renewcommand{\arraystretch}{1.1}
    \scalebox{1}{
    \begin{tabularx}{\textwidth}{ccXSSc}
        \toprule
        & \sithead{Name} & \sithead{Description} & \sithead{Examples} & \sithead{Classes} & \sithead{Labels} \\
        \midrule
        \multirow{2}{*}{\rotatebox{90}{Anim}} & Birds 525 & Scraped dataset of bird images from web search \citep{piosenka2023birds}. & 89885 & 525 & Taxonomic \\
        & BioCLIP-Rare & Subset of species in the IUCN Red List categories: Near Threatened through Extinct in the Wild (\href{https://www.iucnredlist.org/}{iucnredlist.org}). & 12000 & 400 & Taxonomic \\
        \cmidrule(lr){2-6}
        \multirow{2}{*}{\rotatebox{90}{Plt \& Fun}} & Fungi & Expert-labeled images of Danish fungi \citep{picek2022danish}. & 1000 & 25 & Scientific \\
        & DeepWeeds & Weed images collected in situ from eight rangelands across northern Australia \citep{deepweeds}. & 17509 & 9 & Common \\
        \cmidrule(lr){2-6}
        \multirow{2}{*}{\rotatebox{90}{Inse}} & Confounding Species & Dataset evaluating models on challenging visually similar species pairs \citep{chiranjeevi2023deep}. & 100 & 10 & Mixed \\
        & Insects-2 & Mixed common and scientific name classification for insect pests \citep{Wu2019Insect}. & 4080 & 102 & Mixed \\
        \bottomrule
    \end{tabularx}
    }
\end{table*}

\noindent\textbf{Metrics.} We evaluate model performance using top-1 zero-shot accuracy across all benchmark datasets. For datasets containing taxonomic information, we report accuracy based on scientific names, ensuring fine-grained classification. For datasets that lack explicit taxonomic details, we use the category labels as defined by the original benchmark authors. We compute an aggregate performance metric, which represents the weighted average accuracy over all unique class labels across the benchmark suite. This aggregate metric provides an overall view of model performance across diverse tasks.

To account for statistical variability, we include 95\% confidence intervals for all reported metrics, calculated using the binomial proportion confidence interval method (denoted by ±). This provides a robust understanding of the performance and reliability of our results. As suggested during the review process, we incorporated this statistical analysis to strengthen the evaluation of our models.

%We report the Top-1 zero-shot accuracy on the benchmark datasets described in Section \ref{sec:BioTrove-Trainbenchmarks} and Table \ref{tab:eval-data}. The results are based on BioTrove-CLIP trained on the BioTrove-Train subset, detailed in Section \ref{BioTrove-Train}. We evaluate the model using three different checkpoints: BioCLIP \citep{stevens2023bioclip}, OpenCLIP \citep{radford2021learning}, and MetaCLIP-cc \citep{xu2023metaclip}. 

\textbf{Overview of results.} In Table \ref{tab:maintab}, we report the results of our core benchmark suite. At a high level, we observe that \textsc{BioTrove-CLIP} variants achieve the \textbf{best accuracy averaged over benchmarks}. In particular, they perform extremely well on \textsc{BioTrove-Balanced} (a remarkable \textbf{91.1} top-1 accuracy over 2250+ class labels). \textsc{BioTrove-CLIP} also does very well on the Fungi dataset (even though the Fungi class is not central to \textsc{BioTrove-Train}),  and the DeepWeeds dataset. Therefore, \textsc{BioTrove-CLIP} exhibits strong generalization capabilities across diverse datasets.

We also observe that \textsc{BioCLIP} performs very well on \textsc{BioTrove-Unseen} and \textsc{BioCLIP-Rare}. The reasons might be that \textsc{BioCLIP} has seen approximately 450K species, and there might be nontrivial overlap with the species set in \textsc{BioTrove-Unseen}. On the other hand, it could be that \textsc{BioTrove-CLIP} suffers from forgetting issues while training on BioTrove-Train. %In fact, deeper analysis of that benchmark dataset shows significant accuracy improvements (for OpenAI checkpoint base, it is 12.9 for OpenAI to 47.1 for \textsc{BioTrove-CLIP} from OpenAI; similarly, for MetaCLIP-cc, we see a 19\% improvement in accuracy). 
For BioCLIP-Rare, the dataset is a subset from EOL which BioCLIP did not see before, but TreeofLife contains the majority of the EOL dataset. % Notably, for the other two checkpoints, BioTrove-CLIP improves accuracy by at least 11\%.

{\textbf{Limitations}.} We also evaluated all models on the challenging \textsc{Confounding-species} benchmark introduced in \cite{chiranjeevi2023deep}, but find that all models perform at or below random chance and do not report results here; this could be an interesting avenue for follow-up work. 

In Table \ref{tab:taxclass} in the Appendix, we report model performance at different levels of the taxonomic hierarchy. Generally, we find that models trained on web-scraped data perform better with common names, whereas models trained on specialist datasets perform better when using scientific names. Additionally, models trained on web-scraped data excel at classifying at the highest taxonomic level (kingdom), while models begin to benefit from specialist datasets like BioTrove-Train and Tree-of-Life-10M at the lower taxonomic levels (order and species).
%However, \textsc{BioTrove-CLIP} shows a performance decline at taxonomic levels below the species level. This is likely because our training metadata structure allows for classifications solely by referring to species information. 
From a practical standpoint, this is not problematic: \textsc{BioTrove-CLIP} is highly accurate at the species level, and higher-level taxa can be deterministically derived from lower ones.

\begin{table*}[t]
    \caption{\sl \textbf{\textsc{BioTrove-CLIP} performance on various benchmarks.} 
    The top three rows are pre-trained checkpoints: \text{OpenAI-B} refers to OpenAI's ViT-B-16 model, \text{BioCLIP-B} refers to the BioCLIP ViT-B-16 model, and \text{MetaCLIP-L} refers to the MetaCLIP-cc ViT-L-14 model. The bottom three rows are \textsc{Biotrove-Clip} models fine-tuned on different checkpoints: \text{BT-Clip-O} (from OpenAI-B), \text{BT-Clip-B} (from BioCLIP-B), and \text{BT-Clip-M} (from MetaCLIP-L). Benchmark abbreviations: BTU (Biotrove-Unseen, n=300), BTB (Biotrove-Balanced, n=2253), BCR (BioCLIP-Rare, n=400), F (Fungi, n=25), I2 (Insects-2, n=102), B (Birds-525, n=525), LS (Life-Stages, n=20), and DW (DeepWeeds, n=9). 95\% confidence intervals (±) are included.}
    \label{tab:maintab}
    \centering
    \renewcommand{\arraystretch}{1.4} % Adjust row height
    \setlength\tabcolsep{3.1pt} % Adjust column width
    \resizebox{1.0\textwidth}{!}{%
    \begin{tabular}{@{}lccccccccc@{}}
        \toprule
        \textbf{Model} & \textbf{BTU} & \textbf{BTB} & \textbf{BCR} & \textbf{F} & \textbf{I2} & \textbf{B} & \textbf{LS} & \textbf{DW} & \textbf{Weighted Avg.} \\ 
        \midrule
        \textbf{OpenAI-B} & \text{12.9} {\scriptsize± 0.6} & \text{7.3} {\scriptsize± 0.15} & \text{10.9} {\scriptsize± 0.56} & \text{11.5} {\scriptsize± 1.98} & \text{10.2} {\scriptsize± 0.93} & 50.0 {\scriptsize± 0.33} & \textbf{\color{blue}56.5} {\scriptsize± 3.97}  & 10.3 {\scriptsize± 0.45} & 14.7 \\
        \textbf{BioCLIP-B} & \textbf{\color{violet}68.2} {\scriptsize± 0.83} & 62.2 {\scriptsize± 0.28} & \textbf{\color{violet}30.2} {\scriptsize± 0.82} & 45.1 {\scriptsize± 3.08} & \textbf{\color{violet}20.8} {\scriptsize± 1.25} & \textbf{\color{blue}68.7} {\scriptsize± 0.30} & 18.0 {\scriptsize± 3.07} & \textbf{\color{blue}19.9} {\scriptsize± 0.59} & 58.5 \\
        \textbf{MetaCLIP-L} & 24.9 {\scriptsize± 0.77} & 15.4 {\scriptsize± 0.21} & 20.5 {\scriptsize± 0.72} & 24.6 {\scriptsize± 2.67} & 16.1 {\scriptsize± 1.13} & \textbf{\color{violet}70.1} {\scriptsize± 0.30} & \textbf{\color{violet}64.3} {\scriptsize± 3.83} & 14.7 {\scriptsize± 0.52} & 25.0 \\
        \midrule
        \textbf{BT-CLIP-O} & 47.1 {\scriptsize± 0.89} & \textbf{\color{violet}91.1} {\scriptsize± 0.17} & 22.9 {\scriptsize± 0.75} & 43.2 {\scriptsize± 3.07} & 16.5 {\scriptsize± 1.14} & 47.8 {\scriptsize± 0.33} & 28.0 {\scriptsize± 3.59} & 17.0 {\scriptsize± 0.56} & \textbf{\color{violet}70.8} \\
        \textbf{BT-CLIP-B} & \textbf{\color{blue}53.8} {\scriptsize± 0.89} & \textbf{\color{blue}82.2} {\scriptsize± 0.22} & \textbf{\color{blue}23.7} {\scriptsize± 0.76} & \textbf{\color{blue}53.9} {\scriptsize± 3.09} & \textbf{\color{blue}16.9} {\scriptsize± 1.15} & 57.1 {\scriptsize± 0.32} & 15.0 {\scriptsize± 2.86} & 18.4 {\scriptsize± 0.57} & 67.2 \\
        \textbf{BT-CLIP-M} & 44.3 {\scriptsize± 0.89} & \textbf{\color{violet}91.1} {\scriptsize± 0.17} & 21.8 {\scriptsize± 0.74} & \textbf{\color{violet}54.7} {\scriptsize± 3.09} & 5.1 {\scriptsize± 0.68} & 42.5 {\scriptsize± 0.32} & 26.3 {\scriptsize± 3.52} & \textbf{\color{violet}49.9} {\scriptsize± 0.74} & \textbf{\color{blue}69.5} \\
        \bottomrule
    \end{tabular}}
\end{table*}
Addressing these limitations will further enhance the applicability of models like \textsc{BioTrove-CLIP} in real-world biodiversity monitoring tasks.

 \vspace{-0.15in}
\section{Concluding Discussion}
\label{sec:future}
\vspace{-0.1in}
We introduce \textsc{BioTrove}, the largest publicly accessible dataset designed to advance AI for biodiversity applications. This dataset, curated from the iNaturalist community science platform,  includes 161.9 million images, surpassing existing datasets in scale by an order of magnitude. We anticipate that \textsc{BioTrove} will enable the development of AI models that can enable various digital tools ranging from pest control strategies, crop monitoring, and worldwide biodiversity assessment and environmental conservation.

We also believe that \textsc{BioTrove} can be used as a unique testbed for measuring progress on fine-grained image recognition. The success of \textsc{BioTrove-CLIP} on \textsc{BioTrove-Unseen} underscores the importance of scaling up per-category sample size, or vertical scaling~\cite{Feuer2023ExploringDI}, in achieving high accuracy on long-tailed extreme-imbalance classification. 
However, \textsc{BioCLIP} continues to exhibit superior performance on several datasets, and we believe that this is because \textsc{TreeofLife-10M} contains an order-of-magnitude more \emph{classes} (species) than \textsc{BioTrove-Train}. 
We invite the AI community to create new subsets of \textsc{BioTrove} with varying degrees of balance and species diversity and use our tooling to measure model performance against current benchmarks. 
\vspace{-0.1in}
\section*{Acknowledgements}
We acknowledge support from the AI Research Institutes program supported by NSF and USDA-NIFA under AI Institute for Resilient Agriculture, Award No. 2021-67021-35329, and the NAIRR program for computing support.

\newpage

\bibliography{references}

%%%%%%%%%%%%%%%%%%%%%%%%%%%%%%%%%%%%%%%%%%%%%%%%%%%%%%%%%%%%

\newpage

\appendix

\section{Appendix}

\subsection{Background on CLIP and zero-shot classification}

Unlike traditional vision models, CLIP jointly trains an image encoder and a text encoder to predict the correct pairings of a batch of (image, text) examples, leveraging natural language supervision to enhance generalization \citep{radford2021learning}. CLIP's approach allows it to learn from a wide variety of images and their associated textual descriptions, making it more flexible and general compared to standard vision models. This flexibility is crucial for in various domains, including biodiversity monitoring and agriculture. For instance, CLIP models analyze digital plant specimen images, aiding in pre-processing and filtering for further analysis for agriculture purposes \citep{kommineni2023role, li2023rs}. As for biodiversity, WildCLIP and KI-CLIP facilitate wildlife observation and monitoring with high accuracy and effectiveness in data-sparse settings \citep{gabeff2024wildclip, mou2023monitoring}. These examples underscore the importance of developing and utilizing comprehensive datasets to fully leverage the capabilities of CLIP models in advancing biodiversity and agricultural research.

\subsection{The value of taxonomic information}
Taxonomic classification, the hierarchical arrangement of organisms into categories based on shared characteristics, is foundational in biological sciences. Taxonomy underpins various scientific, ecological, and agricultural applications. It allows for precise identification and classification of species, which is fundamental for understanding biodiversity and monitoring ecosystems. For instance, accurate species identification can aid in tracking invasive species, as noted in studies such as \citep{silvestro2022improving}. In agriculture, detailed taxonomic information helps in identifying pests and beneficial species, thereby improving pest control strategies and crop management; supports ecological research by providing insights into species interactions, distribution patterns, and evolutionary relationships \citep{gharaee2024step}; and is essential for policy-making and conservation planning \citep{sen2021combining}.

%Incorporating detailed taxonomic data into model training significantly enhances the accuracy and generalizability of AI models used in biodiversity and ecological studies. Models trained with this data can more accurately classify species and generalize to new, unseen species, making them robust tools for ecological predictions and assessments \citep{borba2021machine}. This is particularly important in studies where the goal is to identify and monitor a wide range of species across different taxonomic levels \citep{stevens2023bioclip}. Detailed taxonomic data also allows models to provide species-level insights, which are crucial for understanding species-specific traits and behaviors and for making informed conservation decisions \citep{gharaee2024step}.

\subsection{Scientific versus common names}
Although we identify the importance and need to include taxonomic information in the dataset for biodiversity, one potential challenge is the fact that this information is mostly in Latin for which text embedding models often exhibit suboptimal performance due to its status as a low-resource language \citep{stringham2020evaluating}. Nonetheless, Latin remains indispensable as it is the standard for representing scientific names and taxonomic classifications. We therefore integrate common names, scientific names, and detailed taxonomic hierarchies. We believe that such an ``all-encompassing'' approach facilitates the learning of relationships between Latin and English terms, thereby improving the models' applicability in scientific contexts \citep{de2024unsupervised, sprugnoli2020building, wijayanti2023learning}. Furthermore, incorporating taxonomic data into the training process significantly enhances the multimodal capabilities of the models, enabling them to associate visual data with taxonomic terminology \citep{mars2022word, chen2021low}.

\subsection{iNaturalist, iNaturalist Open Data}
\label{iNaturalist_Open_Dataset}

iNaturalist is an online social network for sharing biodiversity information and learning about nature. It serves as a crowdsourced species identification system and organism occurrence recording tool. Users from around the world upload images, making the continuously updated dataset valuable for AI applications in biodiversity and research. Each photo includes detailed metadata: copyright status, location, uploader, time, and taxonomic classification. This diversity in image sources makes iNaturalist an excellent dataset for training AI models intended for real-world applications \citep{unger2021inaturalist, niemiller2021addressing, di2021observing, chiavassa2024fair}. Despite its vast and diverse data, iNaturalist is not directly optimized for AI researchers: arranging this data for use in AI models like CLIP is not straightforward. Each photo has its own page on the iNaturalist website, making it difficult to download images along with all the necessary information in a streamlined manner.

The iNaturalist Open Dataset aims to address some of these challenges. It is one of the world’s largest public datasets of photos of living organisms, structured as a "bucket" of images stored using Amazon Web Service's Simple Storage Service (S3). The dataset includes multiple resized versions of each photo, allowing users to download the size most useful to their research. 

Additionally, the dataset provides four tab-separated CSV files representing observations, observers, photos, and \texttt{taxa\_id}.
These files are generated monthly, capturing a snapshot of the continually changing iNaturalist data. The images in the iNaturalist Open Dataset are licensed under either CC0, CC-BY, or CC-BY-NC and are open for public research. Photos with a CC0 license can be attributed as "[observer name or login], no rights reserved (CC0)". Photos with other Creative Commons licenses can be attributed as "© [observer name or login], some rights reserved ([license abbreviation])". 

\subsection{iNaturalist Details}
\label{sec:inaturalist}

Each image in the iNaturalist Open Dataset can be associated with its appropriate metadata through a group of four metadata CSV files, 
%The metadata structure of these files is made for bookkeeping purposes rather than the purpose of machine learning, making it difficult to work with for machine learning research.
\begin{comment}
\noindent\textbf{Data Acquisition}  We use the Large Data Acquisition Workflow Template (LDAWT) to optimize time consumption in data acquisition. LDAWT employs a parallelized distributed download method, utilizing multiple virtual cloud machines for simultaneous asynchronous downloads, and aggregates the data into shared cloud storage. This tool maximizes efficiency by leveraging cloud computation and available bandwidth.
\end{comment}
% \textbf{Zi: }How do you get the well-processed metadata?  mention all the pains you have when use the INAT open data, and how you solve it. Difinetion of well-processed metadata: the taxa info for each level including  taxon rank , the URLs for download.
%\noindent\textbf{Initial Data Sources}  %Zi:Description of the initial data sources (iNaturalist Open Dataset).
%Naturalist Open Dataset provides separate four metadata CSV files for the user, 
representing photos, observations, taxa, and observers. 

%The photos metafile contains the following metadata information as columns for each photo: photo\_uuid, photo\_id, observation\_uuid, observer\_id, extension, license, width, height, and position. 
The photos metadata file contain nine distinct columns of metadata information of each photo. Of these columns, only photo\_id and observation\_uuid are relevant for us. The value of photo\_id is a identifier number used to access individual photos, the photo’s iNaturalist page can be found by constructing a URL in this format: https://www.inaturalist.org/photos/[photo\_id]. The value of observation\_uuid indicates which observation the photo is associated with, it is used to map the photos metadata to the observations metadata.

An observation represents one user submission of a species encounter to the iNaturalist website. One observation can have multiple photos of the same species but never multiple species. The observation metedata file contains eight distinct columns of metadata information on each observation. The columns relevant to us are observation\_uuid, quality grade, and taxon\_id. Each observation is given a unique number identifier indicated by its observation\_uuid. iNaturalist has its own system to determining the quality of an observation and its associated photos, quality\_grade represents this and can range from "Casual", "Research Grade", or "Needs ID". The value taxon\_id indicates the species is represented in the observation, it is used to map the observations metadata to the taxa metadata.

The taxa metadata file contains information about each specific taxon in iNaturalist, it has has six distinct metadata columns. The columns relevant to us are taxon\_id, name, ancestry, and active. Each specific taxon in iNaturalist has a unique identifier number associated with it, this is its taxon\_id. This taxon\_id will map to the scientific name of the taxon which is represented in the name metadata column. Each taxon also has associated with it a taxonomic ancestry, this is represented as a string of taxon\_ids concatenated together with "$\backslash$" like so "48460/1/47115/47584/1051154". The active column indicated whether the taxon is currently in use in iNaturalist.

The observer metadata file comtains information about each user within the iNaturalist site. For the purpose of machine learning research none of its three metadata columns are relevant.

%\noindent\textbf{Metadata Transformation} %Zi: Explanation of how raw data is transformed into BioTrove-specific metadata. 
While the iNaturalist Open Dataset metadata files provide a plethora of interesting information, its structure makes it inherently cumbersome to use for research. To solve this, we aggregate and process the iNaturalist metadata into a concise and streamlined format for easy query and usage.  

First, the respective CSV files are used to populate a SQL database with each CSV file as its own SQL table. A new aggregate SQL table is created that joins the photos, observations, and taxa tables on its relational columns. Only the metadata columns we deemed relevant are kept and the extraneous non-useful metadata columns are discarded. 

One of the difficulties working with the base iNaturalist metadata files is that it does not contain the image URL, information that is critical in image downloads. We include a new column in the aggregated metadata table that explicitly links to the Amazon S3 URL in which the image is hosted.  

The \textsc{BioTrove} metadata file used for model training contains the metadata columns phylum, class, order, family, genus, species, scientific\_name, common\_name for the seven BioTrove categories \textit{Aves}, \textit{Arachnida}, \textit{Insecta}, \textit{Plantae}, \textit{Fungi}, \textit{Mollusca}, and \textit{Reptilia}.
To ensure that only images and metadata from the seven BioTrove categories appear in our final dataset we use the taxa table to find the taxon in our categories then use it in a SQL query on the ancestry column of our aggregated metadata table.

The taxonomic rank columns are also found utilizing the ancestry metadata column. A difficulty in working with the ancestry metadata is present in that there is not a clear indication of what taxonomic rank a taxon id represents the ancestry string. This problem is exacerbated due to the presence of taxonomic ranks and  dsub ranks whose presence is variable across different species. As such, a custom function is applied to each row to dynamically find the rank of each taxon id in the ancestry and then appropriately populate the taxon id to a metadata column of that rank. This process results in all taxonomies rank represented as metadata columns; only phyllum, class, order, family, genus and species are kept in the BioTrove metadata file. 

The scientific name of a species is found using the name metadata column of our aggregated metadata table. The common name of a species is also useful metadata information. Unfortunately, the iNaturalist Open Data metadata files do not contain the common name information of a species. To address this, we curate a lookup table of the common names in our dataset. This is obtained from the iNaturalist Taxonomy DarwinCore Archive, Having obtained the common names for each species, we append it to the BioTrove-specific metadata. 

%The downloaded metadata contains scientific names and taxon ranks but lacks explicit information about the taxonomic details. It includes an ancestry column that can be parsed to obtain the taxonomy, although this process is not straightforward. To address this, we curated a lookup table encompassing taxonomic details such as kingdom, phylum, class, order, family, genus, species, and common names, which were then appended to the BioTrove-specific metadata.

\subsection{Composition of \textsc{BioTrove} and Related Datasets}

In Table~\ref{tab:dataset_comparison}, we compare \textsc{BioTrove} with existing large-scale biodiversity datasets. \textsc{BioTrove} comprises 161.9 million research-grade images, representing approximately 372,966 species, and significantly surpasses other datasets in terms of both diversity and scale.

\begin{table}[!t]
    \centering
    \caption{Comparison of \textsc{BioTrove} with other biodiversity datasets.}
    \label{tab:dataset_comparison}
    \renewcommand{\arraystretch}{1.4}
    \resizebox{\textwidth}{!}{
    \large
    \begin{tabular}{|>{\raggedright}m{2.0cm}|>{\centering\arraybackslash}m{3.4cm}|>{\centering\arraybackslash}m{3.1cm}|>{\centering\arraybackslash}m{3.0cm}|>{\centering\arraybackslash}m{3.2cm}|>{\centering\arraybackslash}m{3.0cm}|>{\centering\arraybackslash}m{3.0cm}|>{\centering\arraybackslash}m{3.0cm}|}
        \hline
        \textbf{Dataset} & \textbf{\textsc{BioTrove}} & \textbf{Wildlife Insights} & \textbf{TreeOfLife} & \textbf{BioScan} & \textbf{iNaturalist 2017~\cite{van2018inaturalist}} & \textbf{iNaturalist 2019~\cite{inat2019_stats,inat2019_review}} & \textbf{GBIF Backbone~\cite{gbif_taxonomy}} \\ \hline
        \textbf{Size} & 161.9 million images & 148.8 million images (52.6M wildlife images) & 10.4M images & 1.1M images & 675,170 images & 13.1M images & 7.5M records \\ \hline
        \textbf{Diversity} & 366.6K species & 3,682 species & 454.1K species & 8.3K species & 5,089 species & 166.8K species & Millions of species \\ \hline
        \textbf{Labels Provided} & Common/scientific names, taxonomic hierarchies & Species, location, timestamps, behavioral tags & Common/scientific names, taxonomic hierarchies & Scientific names, taxonomic ranks (family-species), DNA barcodes & Common/scientific names, taxonomic ranks (genus-species) & Common/scientific names, taxonomic ranks (genus-species) & Species names, OTU identifiers \\ \hline
        \textbf{Data Source} & iNaturalist Open Dataset & Camera traps, sensors & iNaturalist, EOL, BioScan-1M & Malaise trap specimens, DNA-barcodes & iNaturalist & iNaturalist & Catalogue of Life, iBOL, UNITE, WoRMS, etc. \\ \hline
        \textbf{Key Features} & AI-ready pipeline, high-quality annotations, supports \textsc{BioTrove-CLIP} & Automated processing, AI species recognition & Rich hierarchical data, metadata, supports \textsc{BioCLIP} & Insect-focused, high-resolution, taxonomic data, DNA codes & Imbalanced classes, fine-grained taxonomy & Large-scale species data, growth from 2017 & Comprehensive taxonomy, cross-referencing datasets \\ \hline
        \textbf{AI-Ready} & Yes & Yes & Yes & Yes & No & No & No \\ \hline
    \end{tabular}
    }
\end{table}

\subsection{Composition of BioTrove-Train}

\begin{figure}[h]
    \centering
    \includegraphics[width=\textwidth]{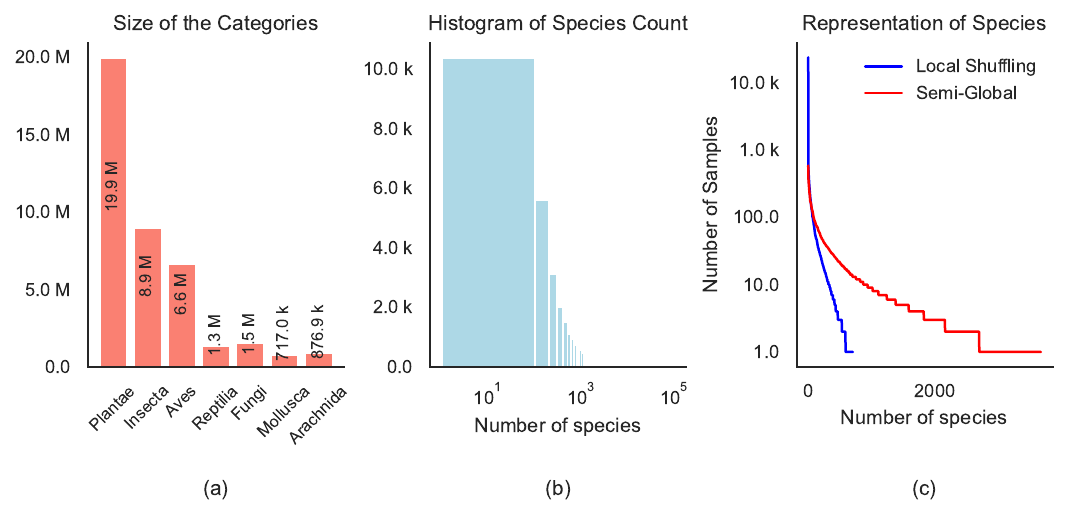}
    \caption {\sl BioTrove-Train Dataset Analysis: a) Consistent category distribution across BioTrove-Train and BioTrove-116M datasets. b) Species exhibit a long-tailed distribution.     c) Impact of local vs. semi-global shuffling on species representation within training minibatches.}
    \label{fig:40M}
\end{figure}

See Figure \ref{fig:40M} and Table \ref{tab:BioTrove-Train-Categories-species-count}.
 
\begin{table}[!h]
    \centering
    \caption{\sl Number of Unique Species in Each Category in BioTrove-Balanced. \label{tab:BioTrove-Train-Categories-species-count} }

    \begin{tabular}{|c|c|}
        \hline
        \textbf{Category} & \textbf{Number of Unique Species} \\
        \hline
        Kingdom: Fungi & 281 \\
        
        Kingdom: Plantae & 500 \\
        
        Phylum: Mollusca & 147 \\
        
        Class: Insecta & 500 \\
        
        Class: Arachnida & 136 \\
        
        Class: Reptilia & 189 \\
        
        Class: Aves & 500 \\
        \hline
    \end{tabular}

\end{table}

\subsection{BioTrove-CLIP training details}
\label{sec:deets}

We use \textsc{BioTrove-Train} to train new CLIP-style foundation models, and then evaluate them on zero-shot image classification tasks. Following the implementation of \citet{stevens2023bioclip}, we utilize a ViT-B/16 architecture initialized from the OpenAI pretrained weights for our main model, and train for 40 epochs. 

In addition, we also train a ViT-L/14 model from the MetaCLIP~\cite{xu2023metaclip} checkpoint for 12 epochs, and a ViT-B/16 from the BioCLIP checkpoint for 8 epochs. We select the AdamW optimizer from \citet{loshchilov2019decoupled} along with a cosine learning rate scheduler, as this has previously been shown to perform well for CLIP pretraining~\citep{DBLP:journals/corr/abs-2103-00020}. We conduct twenty rounds of hyperparameter optimization using Ray Tune~\cite{DBLP:journals/corr/abs-1807-05118} to determine the optimal learning rate, $\beta_1$, $\beta_2$ and weight decay settings. 

We train our models for a combined 10 days on 8xH100 nodes in bfloat16 precision~\cite{DBLP:journals/corr/abs-1905-12322} with gradient checkpointing, computing loss with local features, and utilizing static graph optimization for DDP.

\subsection{Additional BioTrove-CLIP results}

In Table \ref{tab:taxclass}, we report model performance at different levels of the taxonomic hierarchy. Generally, we find that models trained on web-scraped data perform better with common names, whereas models trained on specialist datasets perform better when using scientific names. Additionally, models trained on web-scraped data excel at classifying at the highest taxonomic level (kingdom), while models begin to benefit from specialist datasets like BioTrove-Train and Tree-of-Life-10M at the lower taxonomic levels (order and species).

However, \textsc{BioTrove-CLIP} shows a performance decline at taxonomic levels below the species level. This is likely because our training metadata structure allows for classifications solely by referring to species information. From a practical standpoint, this is not problematic for the species in our test set since \textsc{BioTrove-CLIP} is highly accurate at the species level, and higher-level taxa can be deterministically derived from the lower ones.

Furthermore, the OpenCLIP and MetaCLIP baselines outperform \textsc{BioTrove-CLIP} on the life stages benchmark. This highlights the importance of retaining the general linguistic capabilities of the pretrained CLIP models for hybrid tasks.

\begin{table}[h!]
\centering
\caption{Performance Comparison Across Benchmarks: This table compares the performance of \textsc{BC-iNat21} (trained solely on the iNaturalist 2021 dataset) and \textsc{BT-Clip} (trained from the \textsc{BioCLIP} checkpoint, originally trained on the \textsc{TreeOfLife} dataset). Metrics include Top-1 Accuracy and Top-5 Accuracy.}
\label{tab:taxclass}
\resizebox{\columnwidth}{!}{%
\begin{tabular}{lrrrr}
\toprule
\textbf{Benchmark} & \textbf{BC-iNat21 Top-1} & \textbf{BC-iNat21 Top-5} & \textbf{BT-Clip Top-1} & \textbf{BT-Clip Top-5} \\ 
\midrule
\textsc{BioTrove Unseen} & 0.2100 & 0.3470 & 0.5380 & 0.8220 \\ 
\textit{Fungi} & 0.4420 & 0.7550 & 0.5390 & 0.7590 \\ 
\textsc{Life-Stages} & 0.2867 & 0.8617 & 0.1500 & 0.8600 \\ 
\textsc{DeepWeeds} & 0.2057 & 0.6897 & 0.1840 & 0.5740 \\ 
\textit{Insects-2} & 0.0103 & 0.0483 & 0.1690 & 0.5710 \\ 
\textit{Birds-525} & 0.5030 & 0.6330 & 0.5710 & 0.7540 \\ 
\textsc{BioCLIP-Rare} & 0.1490 & 0.2790 & 0.2370 & 0.7600 \\ 
\textsc{BioTrove Balanced} & 0.5020 & 0.6450 & 0.5180 & 0.6610 \\ 
\bottomrule
\end{tabular}
}
\end{table}

\subsection{Additional BioTrove-CLIP Comparative Analysis}
We conducted a comparative evaluation of the top-1 and top-5 zero-shot accuracy of the \textsc{BioCLIP} model, which was trained exclusively on the iNaturalist 2021 (iNat21) dataset, and the \textsc{BioTrove-CLIP} model, initialized from \textsc{BioCLIP} checkpoints originally trained on the \textsc{TreeOfLife} dataset. The comparison highlights the performance differences across various benchmarks, as presented in Table~\ref{tab:performance_comparison}.

Our analysis shows that models trained on the \textsc{BioTrove} dataset consistently outperform those trained solely on iNat21, particularly in benchmarks such as \textsc{BioTrove-Unseen}, \textit{Fungi}, and \textit{Insects-2}. While certain benchmarks like \textsc{Life-Stages} and \textsc{DeepWeeds} show moderate differences, the results emphasize the advantages of training on \textsc{BioTrove}, leading to enhanced model accuracy and robustness.

The following table provides detailed performance metrics for both models across various benchmarks, comparing their top-1 and top-5 accuracy scores with associated confidence intervals.

\begin{table}[h!]
\centering
\caption{Performance Comparison Across Benchmarks: This table compares the performance of \textsc{BC-iNat21} (trained solely on the iNaturalist 2021 dataset) and \textsc{BT-Clip} (trained from the \textsc{BioCLIP} checkpoint, originally trained on the \textsc{TreeOfLife} dataset). Metrics include Top-1 Accuracy and Top-5 Accuracy. \textsc{BC-iNat21} refers to BioCLIP (iNat21), and \textsc{BT-Clip} refers to BioTrove-CLIP (BioCLIP checkpoint from TreeOfLife).}
\label{tab:performance_comparison}
\begin{tabular}{>{\raggedright\arraybackslash}p{3.5cm} >{\centering\arraybackslash}p{2.2cm} 
>{\centering\arraybackslash}p{2.2cm} 
>{\centering\arraybackslash}p{2.2cm} 
>{\centering\arraybackslash}p{2.2cm}}
\toprule
\textbf{Benchmark} & \textbf{BC-iNat21 Top-1 Acc.} & \textbf{BC-iNat21 Top-5 Acc.} & \textbf{BT-Clip Top-1 Acc.} & \textbf{BT-Clip Top-5 Acc.} \\ 
\midrule
\textsc{BioTrove-Unseen} & 0.2100 & 0.3470 & 0.5380 & 0.8220 \\ 
\textit{Fungi} & 0.4420 & 0.7550 & 0.5390 & 0.7590 \\ 
\textsc{Life-Stages} & 0.2867 & 0.8617 & 0.1500 & 0.8600 \\ 
\textsc{DeepWeeds} & 0.2057 & 0.6897 & 0.1840 & 0.5740 \\ 
\textit{Insects-2} & 0.0103 & 0.0483 & 0.1690 & 0.5710 \\ 
\textit{Birds-525} & 0.5030 & 0.6330 & 0.5710 & 0.7540 \\ 
\textsc{BioCLIP-Rare} & 0.1490 & 0.2790 & 0.2370 & 0.7600 \\ 
\textsc{BioTrove Balanced} & 0.5020 & 0.6450 & 0.5180 & 0.6610 \\ 
\bottomrule
\end{tabular}
\end{table}

As demonstrated, the model trained on \textsc{BioTrove} exhibits superior performance in most categories, particularly when evaluated on rare and unseen species, underscoring the importance of diverse and large-scale datasets like \textsc{BioTrove} for enhancing biodiversity AI models.

\end{document}

\maketitle

\section{Dataset Documentation and Intended Uses}
Arboretum is a 134.6M sample dataset designed to advance AI for biodiversity applications by providing a large-scale, accurately annotated multimodal dataset that includes images and corresponding textual descriptions for a diverse set of species. Arboretum aims to facilitate the development of AI models for species identification, ecological monitoring, and agricultural research. Additionally, we introduce three new benchmark datasets: Arboretum-Unseen, Arboretum-LifeStages, and Arboretum-Balanced.

\section{URLs for Dataset Access}

We provide URLs for dataset access below.

\begin{itemize}[leftmargin=*]
    \item \textbf{Project Website:} \url{https://baskargroup.github.io/Arboretum/}
    \item \textbf{Hosting Dataset:} \url{https://huggingface.co/datasets/ChihHsuan-Yang/Arboretum}
    \item \textbf{Package:} \url{https://pypi.org/project/arbor-process/}
    \item \textbf{Croissant Metadata Record:} The dataset and detailed information about the dataset can be found on Hugging Face: \url{https://huggingface.co/datasets/ChihHsuan-Yang/Arboretum}
\end{itemize}

\section{Author Responsibility Statement}
As the authors of this submission, we affirm that we bear all responsibility in case of any rights violations or ethical issues associated with this work. We confirm that the submitted work is original, and if it includes third-party content, it is used with proper permissions and attributions.

\section{Hosting, Licensing, and Maintenance Plan}
The dataset is hosted on Hugging Face. We ensured that only images licensed under \texttt{CC0}, \texttt{CC-BY}, or \texttt{CC-BY-NC} from iNaturalist Open Data were included, making them available for public research purposes. 

\section{Dataset Format and Accessibility}
The metadata is provided in parquet format, which includes information such as photo ID, scientific name, detailed taxa information, common name, and photo URL. The Arbor-preprocess package facilitates easy access and processing of this data.

\section{Long-term Preservation}
Our dataset will be available for as long as the iNaturalist Open Dataset is maintained. Our pipeline and metadata will continue to function, ensuring long-term accessibility.

\section{Structured Metadata}
The metadata is provided in parquet file format and includes information such as photo ID, scientific name, detailed taxa information, common name, and photo URL.

\section{Persistent Identifier}
\begin{itemize}[leftmargin=*]
    \item \textbf{GitHub Repository:} \url{https://github.com/baskargroup/Arboretum}
    \item \textbf{Package:} \url{https://pypi.org/project/arbor-process/}
\end{itemize}

\bibliographystyle{unsrtnat}